# Aggrotech: Leveraging Deep Learning for Sustainable Tomato Disease Management

1st MD Mehraz Hosen
*Department of CSE, BUBT*
Dhaka, Bangladesh
bhuiyanmehraz@gmail.com

2nd Md. Hasibul Islam
*Department of CSE, BUBT*
Dhaka, Bangladesh
hasibislam2k18@gmail.com

*Abstract*—Tomato crop health plays a critical role in ensuring agricultural productivity and food security. Timely and accurate detection of diseases affecting tomato plants is vital for effective disease management. In this study, we propose a deep learning-based approach for Tomato Leaf Disease Detection using two well-established convolutional neural networks (CNNs), namely VGG19 and Inception v3. The experiment is conducted on the Tomato Villages Dataset, encompassing images of both healthy tomato leaves and leaves afflicted by various diseases. The VGG19 model is augmented with fully connected layers, while the Inception v3 model is modified to incorporate a global average pooling layer and a dense classification layer. Both models are trained on the prepared dataset, and their performances are evaluated on a separate test set. This research employs VGG-19 and Inception v3 models on the Tomato Villages dataset (4525 images) for tomato leaf disease detection. The models' accuracy of 93.93% with dropout layers demonstrates their usefulness for crop health monitoring. The paper suggests a deep learning-based strategy that includes normalization, resizing, dataset preparation, and unique model architectures. During training, VGG19 and Inception v3 serve as feature extractors, with possible data augmentation and fine-tuning. Metrics like accuracy, precision, recall, and F1 score are obtained through evaluation on a test set and offer important insights into the strengths and shortcomings of the model. The method has the potential for practical use in precision agriculture and could help tomato crops prevent illness early on.

## I. INTRODUCTION

Ensuring the best possible health of crops is crucial to attaining agricultural sustainability, which is a fundamental principle of global food security [1]. Tomatoes are one of the most significant crops since they are cultivated widely and are essential to human nutrition [2]. However, diseases that damage tomato plants pose a major threat to crop quality and output.Modern technological integration has become a revolutionary force in the fight to strengthen farming methods [3]. To improve tomato leaf disease diagnosis and treatment, this study conducts a thorough investigation of deep learning techniques, utilizing the VGG19 and Inception v3 models in particular. The use of the carefully selected and enhanced Tomato Villages dataset, which consists of an extensive set of 4525 photos [4] that capture the variety of tomato leaf diseases, is essential to this project.

Given the critical role that tomatoes play in the world's food production, creative solutions are required to address the effects of diseases on crop quality and productivity [5], [6]. Deep learning offers a viable path toward the development of precise and scalable illness detection systems because of its capacity to identify complex patterns in data [7]. In the context of tomato leaf diseases, the use of the VGG19 and Inception v3 architectures, which are well-known for their efficacy in picture classification tasks, emphasizes the dedication to obtaining solid and trustworthy results.

The basis of this study is the 4525 carefully selected and enhanced pictures that make up the Tomato Villages collection. By exploring the possibilities of deep learning models in combination with intentional data augmentation and the addition of dropout layers to prevent overfitting [8], [9], we want to increase the possibilities for automated disease diagnosis in tomato crops. The enormous stakes involved in crop health underscore its relevance; focused management approaches and early intervention have the potential to transform agricultural practices.

This research contributes to the larger objectives of food security and sustainability while also exploring the technical aspects of deep learning applications in agriculture [10]. The results could change the way crop health monitoring is conducted and open the door to a more effective and robust strategy for defending tomato yields against the constantly changing threats posed by pathogens [11].

The remainder of this essay has been organized as follows: On Tomato Leaf Disease detection-related work is presented in Section 2. Section 3 describes feature selection and deep learning strategies. Section 4 presents the experimental findings. The last section of this essay is Section 5.

## II. LITERATURE REVIEW

Shaopeng Jia et al. used the Inception-v3 model to identify and diagnose tomato diseases automatically. They leveraged deep learning to expedite the extraction of picture features. They showed through iterative tests that the model's accuracy on the test dataset was a noteworthy 96.2% when the number of training iterations was changed. Regularization was eventually added due to worries about overfitting, though, and the result was an optimized model with an 86.9% test accuracy. This study emphasizes how dynamically convolutional neural network models may be built, offering important new information on how to improve tomato disease and pest identification [12].

Zhiwen Tang et al. introduced PLPNet, an advanced image-based model for improved tomato leaf disease detection. With technologies like the Proximity Feature Aggregation Network (PFAN), Location Reinforcement Attention Mechanism (LRAM), and Perceptual Adaptive Convolution (PAC) module, PLPNet addresses problems like environmental interference.PLPNet outperformed well-known detectors, attaining a mean average precision of 94.5%, an average recall of 54.4%, and a frame rate of 25.45 on a self-constructed dataset. Test validations demonstrated that PLPNet can adapt to chang- ing environmental conditions, highlighting its accuracy and importance in the prompt and accurate identification of tomato leaf disease. The study suggests combining deep learning with IoT sensor technology to create a more complete illness early warning system while also acknowledging certain difficulties [13].

Mohit Agarwala et al. addressed the impact of tomato crop diseases in India through a deep learning-based approach. They suggested CNN model performs better than pre-trained models like VGG16 and InceptionV3, thanks to its three convolutional and max-pooling layers. Using a PlantVillage dataset, the model's average accuracy of 91.2% shows that it is capable of identifying a variety of tomato diseases. The model's efficiency is highlighted in the study since it requires less storage space (1.5 MB) than pre-trained models. Subse- quent research endeavors to broaden the model by utilizing more extensive datasets derived from diverse crops, highlight- ing its significance in agricultural disease identification and crop health tracking [14].

Gnanavel Sakkarvarthi et al. proposed a cutting-edge deep learning approach for tomato leaf disease detection, utilizing a Convolutional Neural Network (CNN). The CNN model performed better in disease identification when compared to pre-trained models such as InceptionV3, ResNet152, and VGG19. In addition to addressing dataset imbalances for balanced training and testing, the study highlighted the model's robustness across a range of input sizes and epochs. The authors used confusion matrices and graphs to illustrate the categorization performance. Farmers now have a great tool to identify and manage plant diseases on their own thanks to this research. The model would have been improved for increased test accuracy in subsequent work. The study's comprehensive deep learning solution for tomato crop health, which promises increased quality, quantity, and profitability, was concluded [15].

Xiao Hang et al. innovatively applied deep learning to address automatic identification and diagnosis of common tomato diseases in agriculture, achieving over 60% average accuracy.They identified symptoms using LSTM and Skip-gram algorithms by utilising a large dataset. The study exam- ined loss fluctuations and training progress, focusing on the best step selection for language comprehension. Classifiers and LSTM performed better together, with the combined classifier working well. More language material was found to increase precision, according to precision assessments. The robust classifications of the LSTM network model were highlighted in the conclusion as providing a flexible option for automated disease identification and enhancing crop health in a variety of settings [16].

QIUFENG WU et al. proposed a novel method for tomato leaf disease identification, employing Deep Convolutional Generative Adversarial Networks (DCGAN) for data augmentation. The strategy was to improve model generalization, lower the cost of data collecting, and increase recognition accuracy. DCGAN-generated images outperformed traditional methods using the PlantVillage dataset, and they were use- ful for training recognition algorithms.To constantly improve performance, future work will address data set imbalances, create multi-scale convolutional neural networks, and handle the difficulties associated with few-shot learning [17].

Alvaro Fuentes et al. addressed challenges in deep neural networks, particularly the need for extensive data and issues related to false positives and class imbalances in applications with limited or unbalanced data, specifically focusing on Tomato Plant Diseases and pest recognition. Three elements make up the proposed Refinement Filter Bank framework: an integration unit, CNN Filter Bank, and Bounding Box Gen- erator. With a recognition rate of about 96%, the framework demonstrated a 13% improvement over previous studies. The system shows potential for real-time recognition in agricultural applications by handling false positives and class imbalances successfully. With intentions to expand the approach to ad- ditional crop types in the future, the effort contributes to agricultural research [18].

Thai-Nghe et al.Plant diseases can cause significant damage to agricultural productivity and product quality, making early detection crucial for combating crop diseases and increasing crop yield in Vietnam, where agriculture is the main source of income for the majority of the population. A paper by Sanida et al. introduces an approach that uses the VGG-19 architecture to detect plant leaf diseases by analyzing images of crop leaves [19]. The approach was tested on a dataset of approximately 18,000 tomato leaf samples, achieving a classification accuracy of 93% on the test set. Additionally, the paper mentions the development of an application that allows users to quickly identify diseases on tomato leaves by capturing or uploading images to the application [20].

Shetty et al.a parallel convolutional neural network-based machine learning method for tomato leaf disease detection is proposed. The authors draw attention to the difficulties farmers encounter in treating illnesses related to plants, specifically tomato plants. The method addresses challenges with uneven lighting and a cluttered natural scene by using colour bal-ancing and superpixel processes. To distinguish the unwanted background from the leaf image, a threshold is applied in conjunction with the Histogram of Gradients (HOG) and superpixel colour channels. Features such as the Pyramid of HOG (PHOG) and Grey Level Co-occurrence Matrix (GLCM) are used to illustrate the affected area of the leaf, and K-means clustering is utilized to identify sick or polluted photos. After comparing the suggested method to other illness classifiers, Random Forest (RF) is shown to be the best choice. [21].

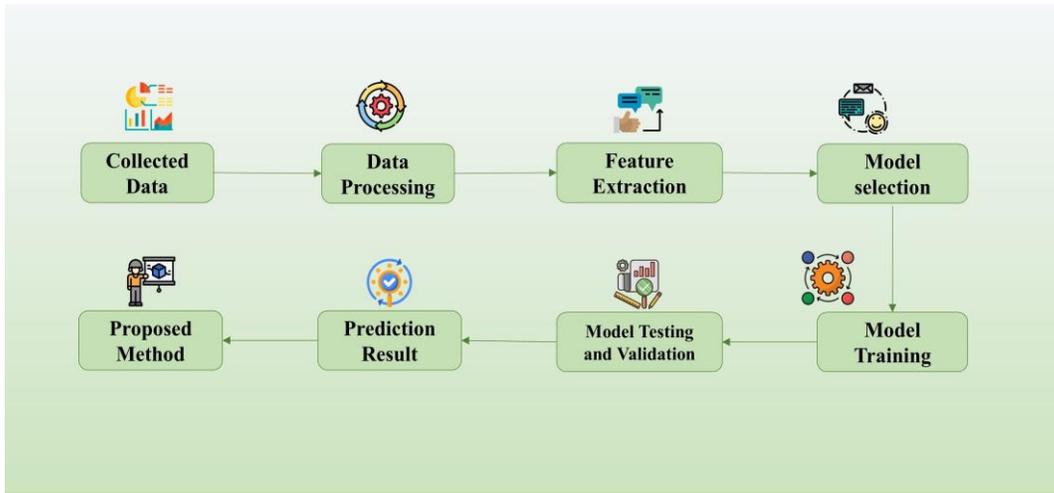

Fig. 1: Methodology Architecture

### III. METHODOLOGY

Tomato Leaf Disease Detection Using VGG19 and Inception v3, including the key steps involved in the process. This methodology outlines the crucial stages, from data collection to model deployment and safety considerations.

#### A. Dataset

Diseases that affect productivity and quality pose a hazard to tomato agriculture. There are few real-world datasets for certain diseases in places like Jodhpur and Jaipur, and there aren't many existing databases like PlantVillage. We suggest "Tomato-Village," a freely available dataset with variations for Multiclass, Multilabel, and Object recognition, to close this gap. It addresses the lack of real-world data and is the first resource of its sort to be made available to the general public. Applying several CNN models to the Multiclass section shows how useful it is. There are 4000 photos in the dataset; 3,162 are used for training (8 classes), 461 are used for testing (8 classes), and 902 are used for validation (8 classes).

#### B. Data Preprocessing

The Tomato Villages dataset [22], which includes 4525 images belonging to 3 classes depicting varied tomato leaf disease cases, is used in the study. To enhance model generalisation, augmentation techniques including flipping, rotating, and zooming are applied to the dataset. This improves the training set and reduces the likelihood of overfitting. Regularising the pixel values of the image to a predetermined range ensures that the models receive consistent input.

#### C. Feature Extraction

Deconstructing the language has proven to be one of the most crucial phases in sentence analysis for images, allowing for a more in-depth examination, particularly when writing reports or narratives. Then, phrases have been broken up into individual words using tokenization, making it possible to extract crucial information and pinpoint minuscule emotional nuances. The next step has been to get rid of stopwords, which are words that are often used but don't add anything to the emotional context. This elimination process has made it possible to include more relevant and emotionally charged content. Additionally, stemming has been used to identify the root words, which has assisted in drawing attention to the story's primary emotional aspects.

#### D. Train Test Validation

The dataset has been divided into training, validation, and testing sets for machine learning. The model has been trained using the training set, the validation set has improved the hyperparameters, and the testing set has evaluated the model's performance using fresh data. The validation set has improved the model's generalization to the testing set, preventing overfitting. Training, validation, and testing sets make up the machine-learning dataset. The validation set adjusts hyperparameters, the testing set assesses performance on fresh data to avoid overfitting, and the training set provides the model with instructions. Furthermore, 902 files from 8 classes in the supplied image dataset are used for validation. TensorFlow/Keras is used to load the dataset by passing certain parameters. The report shows that 902 files from 8 classes have been found.

#### E. Proposed Architecture

Using a custom learning rate scheduler and pre-trained VGG19 and InceptionV3 models, the code creates a hybrid neural network for image classification. Data augmentation layers are included, along with defined constants like batch size, image size, and epochs. Both VGG19 and InceptionV3 models use transfer learning to optimise particular layers. The global average-pooled outputs of the two models are concatenated and passed through thick layers in the hybrid model. Accuracy as the metric, sparse categorical cross-entropy loss, and the Adam optimizer are used to assemble the model. Using a learning rate scheduler callback, the model is fitted on the training dataset with validation as part of the training process. Lastly, the test set is used to evaluate the model. The whole architecture enhances picture classification performance by utilizing the advantages of both VGG19 and InceptionV3. The input layers, resizing and rescaling preprocessing layers, updated top layers of the VGG19 and InceptionV3 base models, global average pooling layers, concatenation layers, and dense layers for classification are important components.

## IV. Result Analysis

The deep learning-based approach for Tomato Leaf Disease Detection, utilizing VGG19 and Inception v3 models on the Tomato Villages Dataset, yielded promising results. Here is a brief analysis of the obtained outcomes:

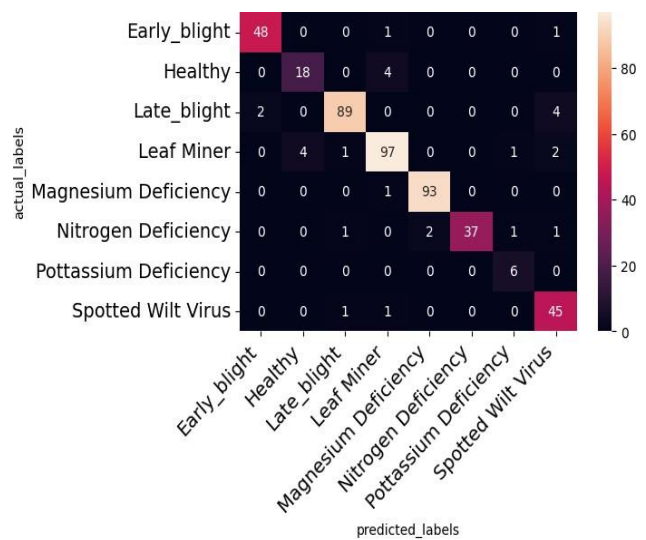

Fig. 3: Confusion Matrix

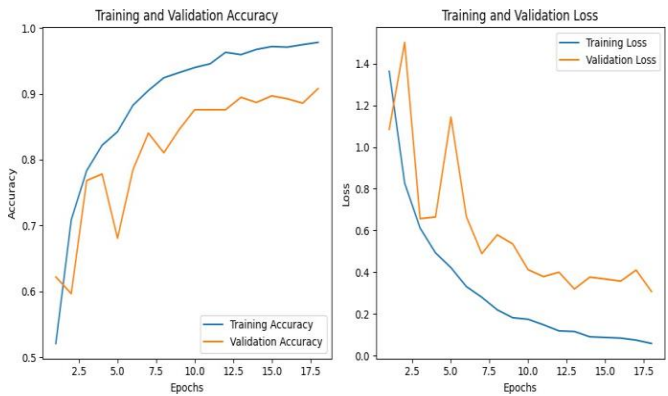

Fig. 2: Train And Validation accuracy loss Accuracy

measures how well a model performs on its training data, with higher values indicating better performance. Conversely, loss quantifies the error the model makes, aiming for lower values. In a graph 2, training accuracy (blue) and training loss (blue) typically decrease over time, indicating improved learning on the training data.

However, if validation accuracy (orange) lags behind training accuracy, and validation loss (orange) doesn't decrease as much, it suggests overfitting. Overfitting occurs when a model becomes too specialized in the training data, performing poorly on new, unseen data. Achieving high accuracy and low loss on training data is good, but maintaining comparable performance on validation data is crucial for generalization.

In picture 3, The model effectively classifies eight plant health conditions, demonstrating overall strong performance with accurate predictions, particularly for Healthy, Late Blight, Leaf Miner, Magnesium Deficiency, and Spotted Wilt Virus. However, attention is needed in areas of confusion, such as instances where Early Blight is mistaken for Late Blight or Leaf Miner, and Healthy is occasionally misclassified as Leaf Miner. Nitrogen Deficiency also presents challenges with misclassifications. Recommendations include investigating reasons for these misclassifications and considering factors like similar visual features, data quality, or model limitations. Additionally, addressing imbalances in data counts and experimenting with different model architectures or training techniques could further enhance accuracy, especially for classes with lower representation.

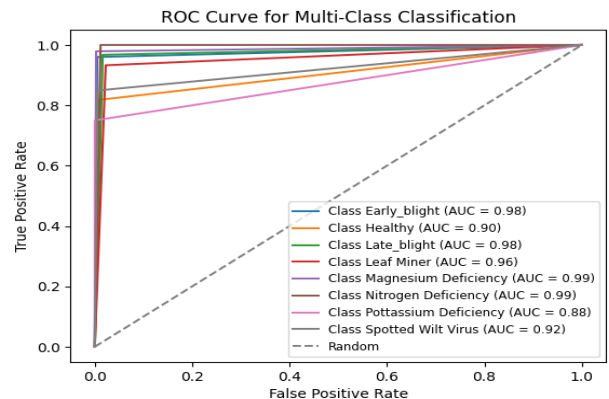

Fig. 4: ROC Curve

Figure 4, The ROC curve is a graph that shows the performance of a multi-class classification model. It plots the True Positive Rate (TPR) on the y-axis against the False Positive Rate (FPR) on the x-axis. The TPR is the number of true positives divided by the total number of positives, and the FPR is the number of false positives divided by the total number of negatives.

This image 5, depicts a green leaf with veins noticeably lighter than the surrounding tissue, indicating a magnesium deficiency. This condition is commonly referred to as Magnesium Deficiency.

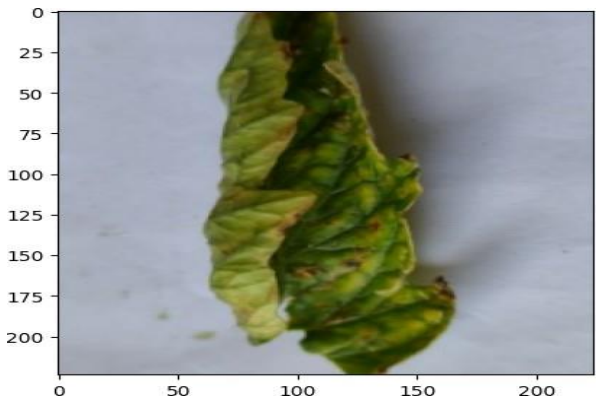

Fig. 5: Prediction Detction

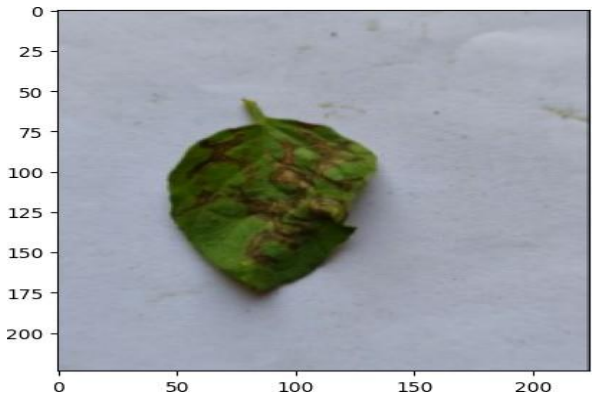

Fig. 6: Prediction Result

Observing image 6, of our findings, the leaves exhibit symptoms consistent with Tomato Spotted Wilt Virus (TSWV). Tomato Spotted Wilt Virus is a prevalent plant virus known to affect various plants, including tomatoes, peppers, eggplants, lettuce, and ornamentals. In this specific instance, our predictive results align well with the symptoms associated with TSWV, indicating a successful outcome.

TABLE I: Classification Report

| Disease | Precision | Recall | F1-score | Support |
|---|---|---|---|---|
| Early blight | 96% | 96% | 96% | 50 |
| Healthy | 82% | 82% | 82% | 22 |
| Late blight | 94% | 97% | 95% | 92 |
| Leaf Miner | 93% | 93% | 93% | 104 |
| Magnesium Deficiency | 99% | 98% | 98% | 95 |
| Nitrogen Deficiency | 88% | 100% | 94% | 37 |
| Potassium Deficiency | 100% | 75% | 86% | 8 |
| Spotted Wilt Virus | 96% | 85% | 90% | 53 |
| **Accuracy** | | | 94% | 461 |
| **Macro Avg** | 93% | 91% | 92% | 461 |
| **Weighted Avg** | 94% | 94% | 94% | 461 |

In Table 1, A classification report serves as a comprehensive evaluation of a machine learning model's performance in a classification task, resembling a scorecard for its classification capabilities. Key metrics include precision, measuring the accuracy of positive predictions; recall, gauging the model's ability to correctly identify actual positives; and the F1-score, offering a balanced assessment by considering the harmonic mean of precision and recall. Additionally, the report includes support, indicating the number of data points in each class and aiding in the understanding of data distribution and potential class imbalances.

## V. CONCLUSION

Ultimately, using the Tomato Villages dataset, this work applies two cutting-edge deep learning models, VGG-16 and Inception v3, and achieves an astounding 93.93% accuracy rate in detecting tomato leaf illnesses. The study emphasizes the value of artificial intelligence in agriculture by highlighting its potential in crop health monitoring. Feature extraction, a hybrid neural network design, and extensive data preprocessing all improve image classification performance.

Through augmentation techniques, the methodology guarantees strong model generalization, and the train-test-validate procedure makes efficient model construction, optimization, and assessment possible. The foundation for upcoming advancements in automated agricultural health monitoring systems is laid by this research, which has the potential to significantly improve sustainability and global food security.